\documentclass{article}

\PassOptionsToPackage{numbers, compress}{natbib}

\usepackage[preprint]{neurips_2025}

\usepackage[utf8]{inputenc} 
\usepackage[T1]{fontenc}    
\usepackage{hyperref}       
\usepackage{url}            
\usepackage{booktabs}       
\usepackage{amsfonts}       
\usepackage{nicefrac}       
\usepackage{microtype}      
\usepackage{multirow}
\usepackage{slashbox}
\usepackage{adjustbox}
\usepackage[table]{xcolor}    
\usepackage[most]{tcolorbox}
\definecolor{block-gray}{gray}{0.9}
\usepackage{subcaption}
\usepackage{enumitem}
\usepackage{subfiles} 
\usepackage{colortbl}
\usepackage{pdfpages}

\title{Towards Generalized Synapse Detection Across Invertebrate Species}

\author{
\textbf{Samia Mohinta}$^{1,2,*\dagger}$ \\
\texttt{sm2667@cam.ac.uk}
\And
\textbf{Daniel Franco-Barranco}$^{1,2,3,*}$ \\
\texttt{dfranco@mrc-lmb.cam.ac.uk}
\And
\textbf{Shi Yan Lee}$^{1,2}$ \\
\texttt{syl58@cam.ac.uk}
\And
\textbf{Albert Cardona}$^{1,2}$ \\
\texttt{ac2040@cam.ac.uk}
\\[1em]
$^{1}$MRC Laboratory of Molecular Biology, University of Cambridge, UK \\
$^{2}$Department of Physiology, Development and Neuroscience, University of Cambridge, UK \\
$^{3}$Donostia International Physics Center (DIPC), San Sebastian, Spain \\
\\
$^{*}$Equal contribution \quad $^{\dagger}$Corresponding author
}

\begin{document}

\maketitle

\begin{abstract}
Behavioural differences across organisms—whether healthy or pathological—are closely tied to the structure of their neural circuits. Yet, the fine-scale synaptic changes that give rise to these variations remain poorly understood, in part due to persistent challenges in detecting synapses reliably and at scale. Volume electron microscopy (EM) offers the resolution required to capture synaptic architecture, but automated detection remains difficult due to sparse annotations, morphological variability, and cross-dataset domain shifts. To address this, we make three key contributions. First, we curate a diverse EM benchmark spanning four datasets across two invertebrate species: adult and larval \textit{Drosophila melanogaster}, and \textit{Megaphragma viggianii} (micro-WASP). Second, we propose \textsc{SimpSyn}, a single-stage Residual U-Net trained to predict dual-channel spherical masks around pre- and post-synaptic sites, designed to prioritize training and inference speeds and annotation efficiency over architectural complexity. Third, we benchmark \textsc{SimpSyn} against Buhmann et al.'s Synful~\cite{Buhmann2021AutomaticSet}, a state-of-the-art multi-task model that jointly infers synaptic pairs. Despite its simplicity, \textsc{SimpSyn} consistently outperforms Synful in F1-score across all volumes for synaptic site detection. While generalization across datasets remains limited, \textsc{SimpSyn} achieves competitive performance when trained on the combined cohort. Finally, ablations reveal that simple post-processing strategies—such as local peak detection and distance-based filtering—yield strong performance without complex test-time heuristics. Taken together, our results suggest that lightweight models, when aligned with task structure, offer a practical and scalable solution for synapse detection in large-scale connectomic pipelines.

\end{abstract}

\section{Introduction}

The brain’s capacity to perceive, learn, and adapt is fundamentally shaped by the organization of its synaptic connections. Synapses—the junctions where neurons communicate—undergo dynamic structural and functional changes in response to development, experience, and disease ~\cite{Winding2023TheBrain, Wolff1992SynapticSystems, Lepeta2016Synaptopathies:Students}. Capturing these fine-grained synaptic patterns is essential for understanding both normal brain function and neurological disorders.

Advances in volume electron microscopy (EM) now allow us to image entire brain regions at nanometer resolution. Yet, despite the anatomical richness in these datasets, reliable synapse detection at scale remains a major bottleneck. Synapses are sparse, morphologically diverse, and context-dependent~\cite{Shen2024EmergingCNS}, varying across species, brain regions, and developmental stages. Although current models can predict millions of synaptic sites, they often yield a high rate of false positives. Conversely, applying conservative filters—such as removing synapses with low connection counts~\cite{dorkenwald2024neuronal} or  based on cleft detection~\cite{Buhmann2021AutomaticSet}—can lead to significant false negatives. These trade-offs underscore a deeper issue: the lack of a unified, generalizable detection framework that can robustly operate across biological and imaging variability without dataset-specific tuning.

Most existing synapse detection methods are developed and validated within a single dataset — trained and tested on anatomically similar volumes. These models often rely on densely annotated voxel-level labels such as synaptic clefts~\cite{Heinrich2018SynapticBrain} or segmented neurites ~\cite{dorkenwald2017automated, Staffler2017SynEMConnectomics}, which are tedious to generate and tied to specific imaging conditions. As a result, their performance degrades significantly when applied to data from different species, brain regions, or acquisition modalities.  Recent methods like Synful~\cite{Buhmann2021AutomaticSet} represent meaningful progress by predicting synaptic pairs from sparse point annotations; however, they still require dataset-specific tuning and come with non-trivial computational overhead, which can limit their generalization to large or heterogeneous datasets.

To address this, we propose \textsc{SimpSyn}: a simple yet effective deep learning model with the promise for generalized synapse detection. \textsc{SimpSyn} forgoes complex multi-task learning or test-time heuristics, instead relies on a single-stage Residual 3D U-Net that predicts spherical masks around pre- and post-synaptic sites. It is trained solely on point annotations and paired with lightweight post-processing to infer synaptic connectivity. This task-aligned design significantly reduces annotation burden and computational cost, while still delivering competitive performance across diverse datasets.

We summarize our main contributions as follows:
\begin{itemize}
    \item \textbf{A cross-species synapse detection benchmark}, spanning 16 volumes (in 4 datasets; Figure \ref{fig:data_overview}a)  across adult and larval \textit{Drosophila melanogaster} and \textit{Megaphragma viggianii} (micro-wasp), including a new manually annotated larval fly dataset (Octo).
    \item \textbf{\textsc{SimpSyn}}, a lightweight, task-aligned model for synapse detection that uses only point annotations and requires no manual or per-dataset test-time tuning.
    \item \textbf{Empirical evidence} that \textsc{SimpSyn} outperforms or matches Synful in F1-score across all datasets, both in-distribution and out-of-distribution, despite being 15$\times$ smaller.
    \item \textbf{Ablation studies} demonstrating that simple post-processing techniques (e.g., local peak detection, distance-based filtering) are sufficient to yield strong performance—highlighting that model simplicity need not come at the cost of accuracy.
\end{itemize}

\section{Related work}

To situate our work within the broader landscape of synapse detection approaches in EM, we first examine the morphological and imaging challenges that underlie synapse detection across species, before reviewing recent methodological advances and limitations in automated detection.

\subsection{Synapse morphology across species}
\label{sec:morph_species}
Vertebrate synapses are reliably identified in transmission EM by distinct ultrastructural features—synaptic clefts, vesicle clusters, and active zones \cite{Jagadeesh2014SynapseMicrographs}. These are well-resolved in anisotropic datasets such as MICrONS \cite{Bae2021FunctionalCortex, Kasthuri2015SaturatedNeocortex, Shapson-Coe2021ACortex}, which trade z-resolution for high x–y coverage ($\leq$4 nm/pixel). Combined with predominantly monadic (1-to-1) connectivity, this enables straightforward partner assignment and supports supervised detection pipelines.

Invertebrate systems present greater challenges. Data are typically acquired via FIB-SEM (Focused Ion Beam Scanning Electron Microscope), which provides isotropic resolution (e.g., 8×8×8 nm) but reduces x–y clarity, obscuring key features like synaptic clefts and PSDs. These structures are often faint or noisy, complicating both manual labelling and model training.

\begin{figure*}[h]
\centering
 \includegraphics[width=1\textwidth]{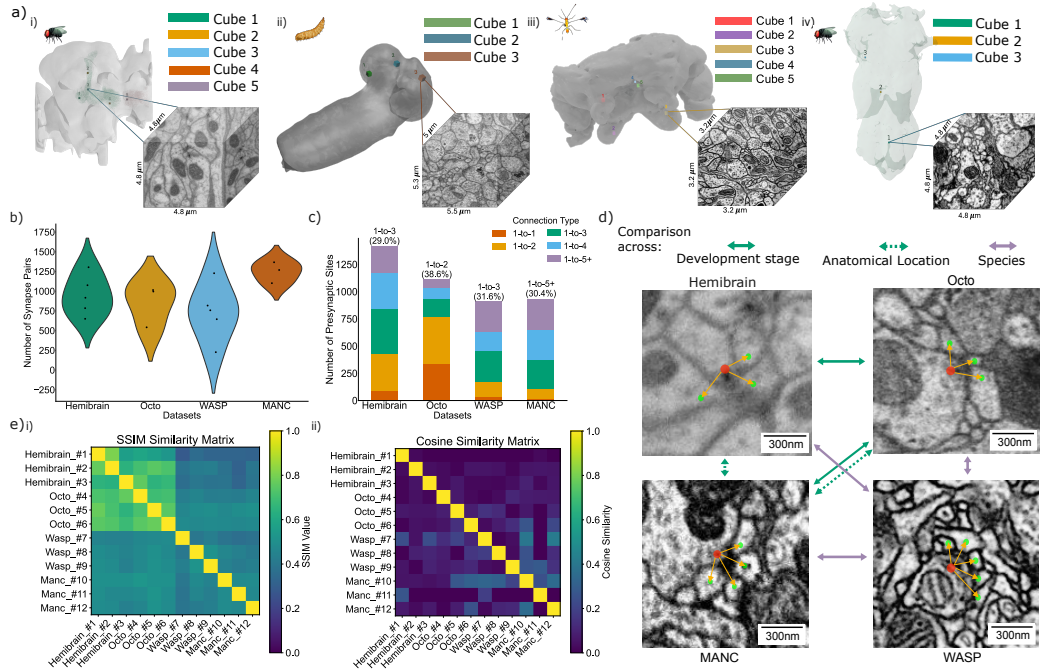}
\caption{\textbf{Diversity in EM structure and synaptic organization across species.} 
a) Spatial distribution of annotated sub-volumes in Hemibrain, Octo, WASP, and MANC with example EM patches showing ultrastructural variability. 
b) Synaptic pair counts per sub-volume. 
c) Presynaptic connection types (1-to-1 to 1-to-5), highlighting Octo’s preference for low-degree connections versus polyadic dominance in others. 
d) Representative polyadic synapses across datasets, organized by developmental stage, region, and species. 
e) Structural similarity analysis. Higher values (yellow) denote greater morphological similarity: i) SSIM reveals higher intra (in-distribution)-/inter(out-of-distribution)-dataset consistency in Hemibrain/Octo and ii) cosine similarity suggests moderate overlaps in WASP and MANC. Scale bars: 300 nm.}
 \label{fig:data_overview}
\end{figure*}

Insect synapses are structurally more complex, with polyadic connections where a single T-bar contacts multiple postsynaptic partners. As shown in Figure~\ref{fig:data_overview}, this complexity varies across FIB-SEM datasets (Hemibrain, Octo, WASP, MANC), developmental stages, and brain regions. WASP and MANC exhibit more frequent 1-to-5+ connections and greater EM variability (Figures~\ref{fig:data_overview}b–e), complicating consistent detection. Structural similarity analyses (Figure~\ref{fig:data_overview}e) reveal larger intra- and inter-dataset shifts. While SSIM captures these morphological changes; cosine similarity is more sensitive to intensity, reflecting contrast-driven resemblance. Both metrics are computed from synaptic patches ($32^3$), underscoring anatomical variability that can hinder model generalization.

\subsection{Synapse labeling and recent approaches}


Automated synapse detection in volume EM has historically been tied to dense, voxel-level annotations—including neurite skeletons~\cite{dorkenwald2017automated}, full segmentations~\cite{Staffler2017SynEMConnectomics,park2022automated,svara2022automated}, clefts~\cite{Heinrich2018SynapticBrain,Bae2021FunctionalCortex, turner2020synaptic,schneider2025inhibitory,svara2022automated}, vesicles~\cite{svara2022automated}, and finely traced membranes~\cite{xiao2018effective}. For example, SynEM~\cite{Staffler2017SynEMConnectomics} requires full neurite segmentation before classifying interfaces as synaptic, while Heinrich et al.~\cite{Heinrich2018SynapticBrain} regress signed-distance maps to cleft masks using a 3D U-Net. These pipelines demonstrate strong performance within their target domains but are tightly coupled to imaging resolution and annotation schemas. As a result, they require extensive re-annotation or tuning to adapt across datasets, limiting their scalability and generalization in comparative or cross-species connectomics.

More recent work has reduced the annotation burden by learning from sparse synaptic labels. In particular, Buhmann et al.\cite{Buhmann2018SynapticBrains}, formulated synaptic partner prediction as an edge-classification task: a 3D U-Net is trained on long-range `edges' linking pre- and post-synaptic neurons, using only point annotations at presynaptic and postsynaptic sites. Overall, the field is increasingly shifting towards the use of this type of annotation for synapse detection, as it is significantly more cost-effective~\cite{Buhmann2018SynapticBrains,dorkenwald2024neuronal,li2024waspsyn,chen2024domain}. 

Buhmann et al.'s Synful approach~\cite{Buhmann2021AutomaticSet} is widely recognized as the state-of-the-art method for synapse detection in large-scale invertebrate EM datasets. In recent years, several landmark projects have relied on Synful for synapse identification, including the reconstruction motor control circuits~\cite{phelps2021reconstruction}, and the ventral nerve cord~\cite{azevedo2024connectomic} of \textit{Drosophila melanogaster}, as well as the complete connectome of the adult fruit fly~\cite{dorkenwald2024neuronal, schlegel2024whole, garner2024connectomic}. Additionally, Synful has been adapted to map the feedforward connectivity within the mouse cerebellar cortex~\cite{nguyen2023structured}.

\subsection{Generalization challenge}
Synapse detection is inherently a challenging problem in computational neuroscience, as noted by Buhmann et al.\cite{Buhmann2018SynapticBrains}. As discussed in \ref{sec:morph_species}, interspecies variability adds another layer of complexity to this problem. Many approaches that report high performance on synapse detection typically evaluate their models on data drawn from the same domain as the training set. Even when applied to different brain regions, these methods often assume consistency in staining protocols and tissue characteristics~\cite{Staffler2017SynEMConnectomics, dorkenwald2017automated, park2022automated, xiao2018effective, huang2018fully}. Existing methods frequently rely on the manual tuning of multiple threshold parameters~\cite{xiao2018effective}, or report optimal performances selected from precision-recall curves—both of which presuppose the availability of ground truth annotations. This reliance poses a significant barrier to the automated deployment of these models across diverse datasets, especially in scenarios where such annotations are unavailable. There is a general lack of guidance within the field regarding how existing synapse detection approaches can be effectively adapted to novel datasets.

In response to these challenges, the WASPSYN23 competition~\cite{li2024waspsyn} was recently introduced to foster the development of deep learning methodologies capable of generalizing across domains, with a particular emphasis on advancing synapse segmentation under varying biological and imaging conditions. Taking inspiration, our work avoids reliance on manually optimized thresholds and extreme hyperparameter tuning, and is tested on four different datasets across 16 non-overlapping brain regions. We introduce automated strategies for threshold selection, enabling the model to generalize to previously unseen data without the need for dataset-specific adjustments.

\section{Materials and Methods}

\subsection{Dataset preparation}


To rigorously evaluate our synapse detection methods, we use four heterogeneous datasets spanning two species: three from \textit{Drosophila melanogaster} (adult: Hemibrain and MANC; larval: Octo in this study) and one from \textit{Megaphragma viggianii} (WASP). Except for Octo, all datasets are publicly available. Curated sub-volumes from public datasets contain high-confidence annotations, primarily derived from machine predictions. In contrast, all synapse annotations in Octo were manually labelled.



\textbf{Hemibrain}
The Hemibrain dataset~\cite{scheffer2020connectome} is a FIB-SEM volume of the adult \textit{Drosophila melanogaster} central brain at $8\times8\times8$ nm resolution, featuring $\sim$25,000 neurons and 20M+ synapses. We extract five $600^3$ voxel subvolumes ($4.8\ \mu$m per side); four are used for training, one for testing. 

\textbf{Octo}
Octo is a FIB-SEM volume of the L1 larval \textit{Drosophila melanogaster} central nervous system (brain $+$ VNC) at $8\times8\times8$ nm resolution. Not yet public, it includes $\sim$2,500 manually annotated synapses across three non-overlapping $25\ \mu$m$^3$ sub-volumes from distinct brain regions, where two are used for training and one for testing.

\textbf{MANC}
The MANC dataset~\cite{Takemura2024ACord} captures the ventral nerve cord of a 5-day-old adult male \textit{Drosophila melanogaster} at $8\times8\times8$ nm resolution, with $\sim$10M presynaptic (T-bars) and 74M postsynaptic sites annotated across $\sim$23,000 neurons. We use three $600^3$ voxel sub-volumes with two for training and one for testing. 

\textbf{WASP}
The dataset of the WASPSYN23 challenge~\cite{li2024waspsyn} consists of 14 FIB-SEM volumes (each $416^3$ voxels, $8\times8\times8$ nm) from \textit{Megaphragma viggianii}, covering diverse brain regions across three specimens. We select five volumes from one brain (specimen 3), and use four for training and one for testing.






\subsection{Methods}

\subsubsection{\textsc{SimpSyn}}
\label{sec:naive_net}
\begin{figure*}[h]
\centering
 \includegraphics[width=1\textwidth]{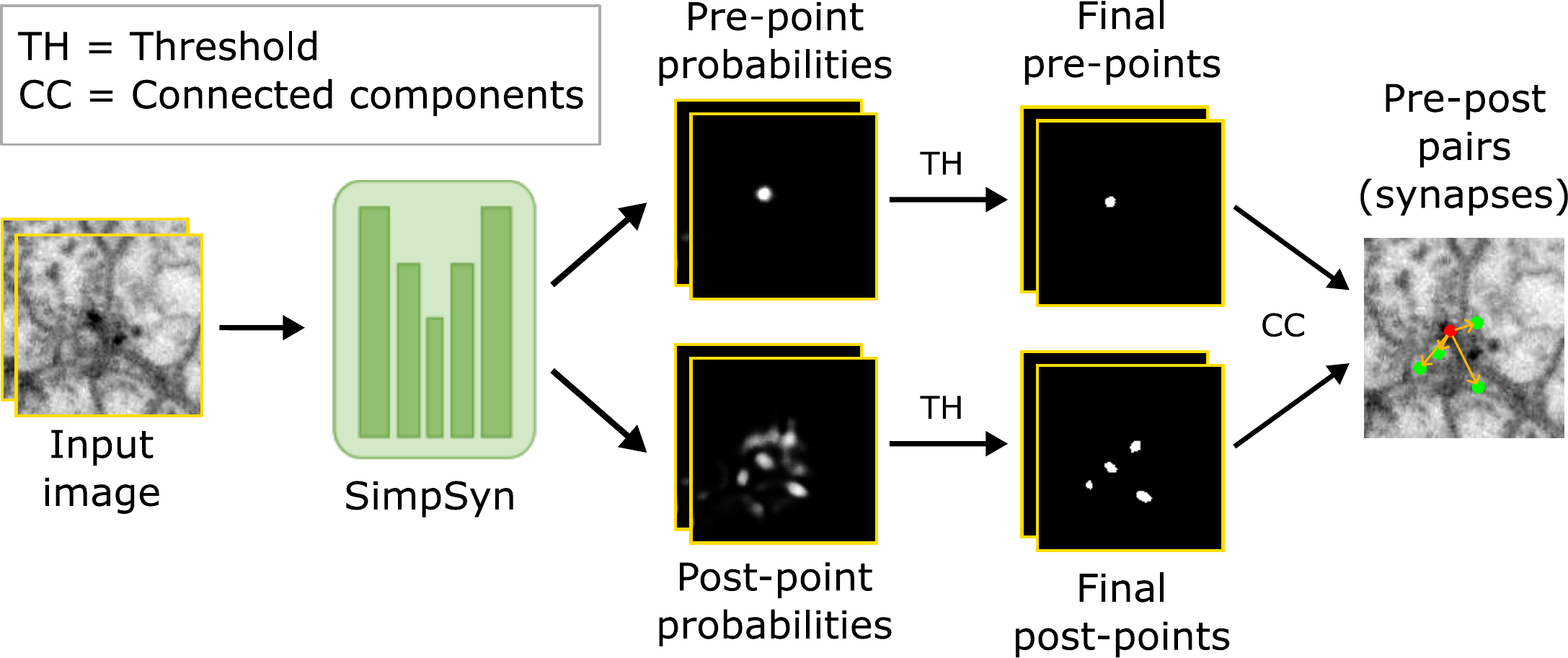}
 \caption{\textbf{General overview of \textsc{SimpSyn} (our Simple Synapse Detection model)}. \textsc{SimpSyn} is a  3D Residual U-Net trained to predict 3D spherical masks of the pre and post synaptic sites. The segmented pre-post masks are paired to form pre-post synapses by using connected components and nearest neighbour pairing.}
 \label{fig:model_overview}
\end{figure*}

\textsc{SimpSyn} is a 3D Residual U-Net~\cite{franco2022stable} for synapse detection that predicts two output channels: one corresponding to pre-synaptic regions and the other to post-synaptic regions as done in~\cite{li2024waspsyn,chen2024domain}. To achieve this, a spherical 3D region is generated and centered at the coordinates of the pre-synaptic and post-synaptic points, following the work by Zhou et. al.~\cite{zhou2019objects} and in detection workflow in Franco-Barranco et al.~\cite{franco2025biapy}. These output masks are subsequently processed using connected component labelling to isolate individual synaptic structures. To establish correspondence between pre-synaptic and post-synaptic sites, each post-synaptic component is paired with its nearest pre-synaptic counterpart based on the nearest neighbour criterion. Figure \ref{fig:model_overview} illustrates the complete pipeline of the \textsc{SimpSyn} model.

\subsubsection{Synful}
\label{sec: synful}
Synful~\cite{Buhmann2021AutomaticSet} is a 3D convolutional neural network designed for synapse detection from sparsely annotated pre- and postsynaptic sites. Synapses are represented as directed voxel pairs $(s, t) \in \Omega^2$, with $s$ and $t$ denoting presynaptic and postsynaptic voxels, respectively. This formulation supports polyadic synapses by modelling each connection independently. We use the MT1 architecture, comprising a single 3D U-Net with shared encoder-decoder pathways and dual output heads for jointly predicting postsynaptic masks and 3D direction vector fields, as illustrated in the Figure S2. At inference time, postsynaptic candidates are identified by thresholding the predicted mask and grouping connected components. Final postsynaptic locations $\hat{t}$ are selected using the peak of a Euclidean distance transform over each component, and corresponding presynaptic sites are computed as $\hat{s} = \hat{t} + \hat{d}(\hat{t})$, forming the predicted pair $(\hat{s}, \hat{t})$.

\section{Experimental setup}



\subsection{Implementation details}

\subsubsection{ \textsc{SimpSyn}}

Our \textsc{SimpSyn} approach is fully integrated within an established deep-learning platform for bio-image analysis, Biapy~\cite{franco2025biapy}, and is implemented using Pytorch version $2.4$. During training, we employ standard data augmentation techniques for EM data~\cite{CiresanNeuralStacks, lee2017superhuman}, including flips, random rotations, brightness and contrast adjustments, and elastic deformations. We set aside a randomly selected $10\%$ of the training synaptic sites for validation, and this split is held constant across all random seeds for a given model configuration. \textsc{SimpSyn} is trained using cross-entropy loss with foreground-background re-weighting based on pixel counts, a cosine-decay learning rate scheduler with warm-up~\cite{loshchilov2016sgdr}, and the AdamW optimizer. Training is performed for $1500$ epochs on eight RTX $3090$ 24GB GPUs for $\sim15$ hours, with early stopping using a patience of $250$ epochs. Inference takes $~\sim5$ minutes. See Table~S3 for a detailed overview of the hyperparameter search.

\subsubsection{Synful}

Our Synful baseline is based on the publicly available source code, which we refactored and encapsulated within a Dockerized environment for reproducibility~\cite{Mohinta2022Catena:Connectomics}. Synful natively relies on TensorFlow version $1.x$ and trained on $128^3$ voxel cubes for $300{,}000$ iterations. Validation is performed as a post-hoc process to select the checkpoint yielding the best performance, in order to reduce computational cost and time of training. The loss function combines binary cross-entropy (for postsynaptic masks) and mean squared error (for presynaptic direction vectors), and is optimized using Adam. We apply data augmentation strategies similar to those in Buhmann et al.~\cite{Buhmann2021AutomaticSet}, including random rotations, intensity scaling, and elastic deformations. 
Training and inference are conducted on a single 12GB Titan XP GPU, with training taking approximately $36$ hours and inference requiring $\sim$12 minutes per $5\mu\mathrm{m} \times 5\mu\mathrm{m} \times 5\mu\mathrm{m}$ volume.


\subsection{Evaluation metrics}

We adopt the evaluation metrics used in the WASPSYN23 challenge~\cite{li2024waspsyn}, where synapse detection accuracy is assessed via bipartite matching to minimize Euclidean distance between predicted and ground-truth synapses~\cite{crouse2016implementing}. Let $D$ and $G$ denote the sets of detected and ground-truth synapses, respectively. A matching $f: D \rightarrow G$ is computed using the Hungarian algorithm to minimize:

\begin{equation*}
\sum_{d \in D} C(d, f(d)), \tag{1}
\end{equation*}

where $C(\cdot)$ is the Euclidean distance between matched pairs. A detection is considered a true positive if the distance is below a $120$-voxel threshold (corresponding to $8\times8\times8$ nm resolution). The F1-score is then computed as:

\begin{equation*}
F_{1} = \frac{2~TP}{2TP + FP + FN}, \tag{2}
\end{equation*}
where $TP$, $FP$, and $FN$ refer to the number of true positives, false positives, and false negatives, respectively. The F1-score thus captures the harmonic mean of precision and recall, offering a balanced measure of detection accuracy. Separate scores are reported for pre- and post-synaptic sites in Table~\ref{tab:result_comparison}.




\subsection{Results and Discussion}

\begin{table}[h!]
\caption{\textbf{Pre- and post-synaptic site detection results} measured using F1 metric (higher is better). The upper row denotes the test datasets, while the first column indicates the training dataset. Consequently, diagonal entries (in \colorbox{block-gray}{gray}) correspond to in-distribution (ID) evaluations, whereas off-diagonal entries represent out-of-distribution (OOD) scenarios. Within each cell, the upper value refers to performance achieved by \textsc{SimpSyn}, and the lower value to that of Synful. For each column, the highest score is highlighted in \textbf{bold}, and the second-highest is \underline{underlined}. \textsc{SimpSyn} consistently achieves the highest F1-scores across all volumes. In nearly all cases, models trained on combined datasets outperform those trained solely on dataset-specific ground truth, highlighting the benefits of cross-domain learning.}
\vspace{0.2cm}
\centering
\begin{adjustbox}{max width=\textwidth}
\begin{tabular}{c|ccccccccccc}
\multicolumn{1}{l}{}           & \multicolumn{5}{c}{Pre synaptic sites}  &   
\multicolumn{6}{c}{Post synaptic sites} \\ \cmidrule{1-6} \cmidrule{8-12} 
\backslashbox{Train}{Test}          & \makebox{Hemibrain}                                             & \makebox{Octo}                                                  & \makebox{WASP}                                                  & \makebox{MANC}                                                  & \makebox{All}                                                   &  & \makebox{Hemibrain}                                             & \makebox{Octo}                                                  & \makebox{WASP}                                                  & \makebox{MANC}                                                  & \makebox{All}                                                   \\ \cmidrule{1-6} \cmidrule{8-12} 
Hemibrain & \begin{tabular}[c]{@{}c@{}}\cellcolor{gray!15}\underline{0.783}\\ \cellcolor{gray!15}0.262\end{tabular} & \begin{tabular}[c]{@{}c@{}}0.021\\ 0.042\end{tabular} & \begin{tabular}[c]{@{}c@{}}0.000\\ 0.000\end{tabular} & \begin{tabular}[c]{@{}c@{}}0.007\\ 0.009\end{tabular} & \begin{tabular}[c]{@{}c@{}}0.203\\ 0.078\end{tabular} &  & \begin{tabular}[c]{@{}c@{}}\cellcolor{gray!15}\underline{0.606}\\ \cellcolor{gray!15}0.437\end{tabular} & \begin{tabular}[c]{@{}c@{}}0.030\\ 0.025\end{tabular} & \begin{tabular}[c]{@{}c@{}}0.000\\ 0.000\end{tabular} & \begin{tabular}[c]{@{}c@{}}0.029\\ 0.013\end{tabular} & \begin{tabular}[c]{@{}c@{}}0.166\\ 0.119\end{tabular} \\
\arrayrulecolor{gray!30}\hline

Octo      & \begin{tabular}[c]{@{}c@{}}0.523\\ 0.017\end{tabular} & \begin{tabular}[c]{@{}c@{}}\cellcolor{gray!15}\underline{0.538}\\ \cellcolor{gray!15}0.150\end{tabular} & \begin{tabular}[c]{@{}c@{}}0.036\\ 0.012\end{tabular} & \begin{tabular}[c]{@{}c@{}}0.156\\ 0.021\end{tabular} & \begin{tabular}[c]{@{}c@{}}0.313\\ 0.046\end{tabular} &  & \begin{tabular}[c]{@{}c@{}}0.139\\ 0.004\end{tabular} & \begin{tabular}[c]{@{}c@{}}\cellcolor{gray!15}\underline{0.425}\\ \cellcolor{gray!15}0.171\end{tabular} & \begin{tabular}[c]{@{}c@{}}0.005\\ 0.002\end{tabular} & \begin{tabular}[c]{@{}c@{}}0.149\\ 0.022\end{tabular} & \begin{tabular}[c]{@{}c@{}}0.179\\ 0.050\end{tabular} \\
\arrayrulecolor{gray!30}\hline

WASP      & \begin{tabular}[c]{@{}c@{}}0.057\\ 0.000\end{tabular} & \begin{tabular}[c]{@{}c@{}}0.400\\ 0.000\end{tabular} & \begin{tabular}[c]{@{}c@{}}\cellcolor{gray!15}\underline{0.781}\\ \cellcolor{gray!15}0.292\end{tabular} & \begin{tabular}[c]{@{}c@{}}0.194\\ 0.148\end{tabular} & \begin{tabular}[c]{@{}c@{}}0.358\\ 0.110\end{tabular} &  & \begin{tabular}[c]{@{}c@{}}0.000\\ 0.000\end{tabular} & \begin{tabular}[c]{@{}c@{}}0.182\\ 0.000\end{tabular} & \begin{tabular}[c]{@{}c@{}}\cellcolor{gray!15}\underline{0.655}\\ \cellcolor{gray!15}0.612\end{tabular} & \begin{tabular}[c]{@{}c@{}}0.300\\ 0.321\end{tabular} & \begin{tabular}[c]{@{}c@{}}0.284\\ 0.233\end{tabular} \\
\arrayrulecolor{gray!30}\hline

MANC      & \begin{tabular}[c]{@{}c@{}}0.647\\ 0.009\end{tabular} & \begin{tabular}[c]{@{}c@{}}0.290\\ 0.005\end{tabular} & \begin{tabular}[c]{@{}c@{}}0.000\\ 0.000\end{tabular} & \begin{tabular}[c]{@{}c@{}}\cellcolor{gray!15}\underline{0.645}\\ \cellcolor{gray!15}0.201\end{tabular} & \begin{tabular}[c]{@{}c@{}}\underline{0.395}\\ 0.054\end{tabular} &  & \begin{tabular}[c]{@{}c@{}}0.028\\ 0.018\end{tabular} & \begin{tabular}[c]{@{}c@{}}0.300\\ 0.002\end{tabular} & \begin{tabular}[c]{@{}c@{}}0.000\\ 0.000\end{tabular} & \begin{tabular}[c]{@{}c@{}}\cellcolor{gray!15}\textbf{0.478}\\ \cellcolor{gray!15}0.364\end{tabular} & \begin{tabular}[c]{@{}c@{}}0.202\\ 0.096\end{tabular} \\
\arrayrulecolor{gray!50}\hline

All       & \begin{tabular}[c]{@{}c@{}}\textbf{0.846}\\ 0.348\end{tabular} & \begin{tabular}[c]{@{}c@{}}\textbf{0.654}\\ 0.320\end{tabular} & \begin{tabular}[c]{@{}c@{}}\textbf{0.810}\\ 0.391\end{tabular} & \begin{tabular}[c]{@{}c@{}}\textbf{0.736}\\ 0.349\end{tabular} & \begin{tabular}[c]{@{}c@{}}\cellcolor{gray!15}\textbf{0.762}\\ \cellcolor{gray!15}0.352\end{tabular} &  & \begin{tabular}[c]{@{}c@{}}\textbf{0.635}\\ 0.248\end{tabular} & \begin{tabular}[c]{@{}c@{}}\textbf{0.506}\\ 0.343\end{tabular} & \begin{tabular}[c]{@{}c@{}}\textbf{0.664}\\ 0.582\end{tabular} & \begin{tabular}[c]{@{}c@{}}\underline{0.475}\\ 0.418\end{tabular} & \begin{tabular}[c]{@{}c@{}}\cellcolor{gray!15}\textbf{0.570}\\ \cellcolor{gray!15}\underline{0.398}\end{tabular}  \\ 
\arrayrulecolor{black}\hline
\end{tabular}
\end{adjustbox}
\label{tab:result_comparison}
\end{table}
\vspace{-1em}

\begin{figure*}[ht]
\centering
 \includegraphics[width=1\textwidth]{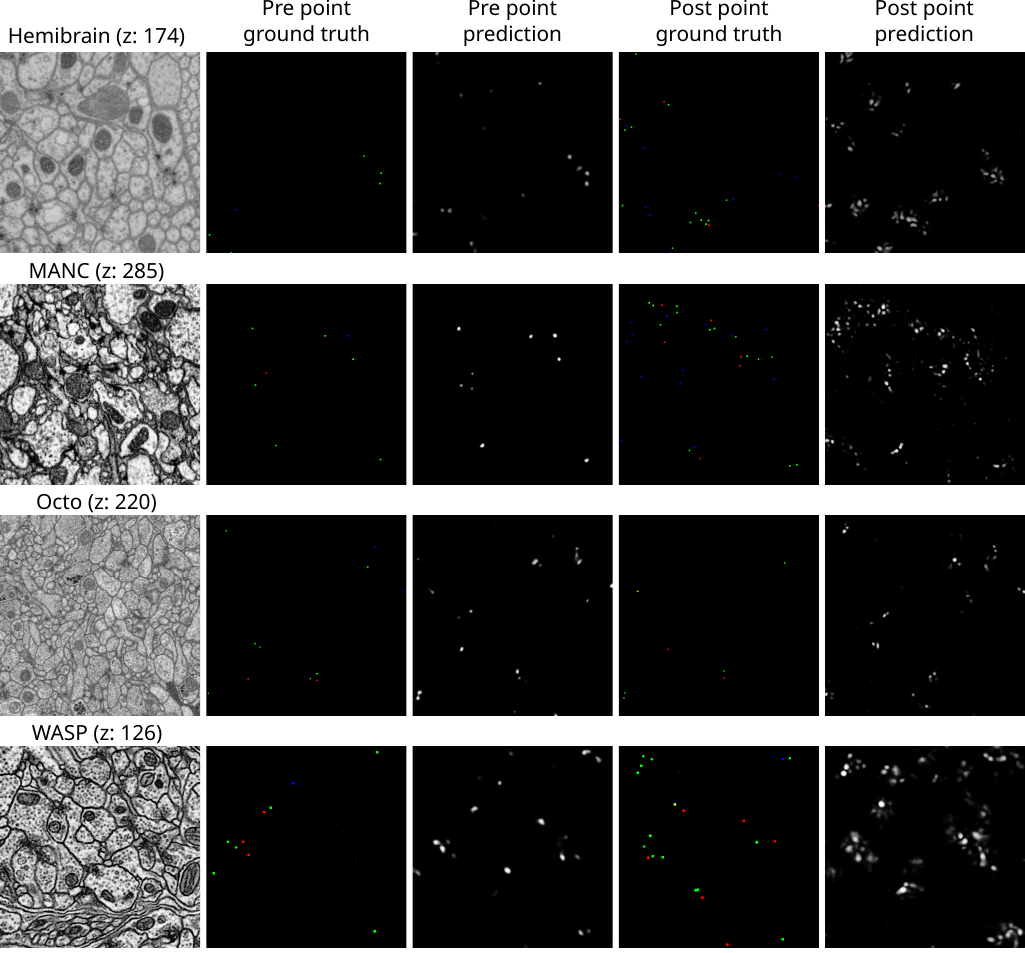}
 \caption{\textbf{Qualitative results \textsc{SimpSyn} on test samples}. Correctly predicted point (true positives) are marked in \textcolor{green}{green},  un-detected points (false negatives) in \textcolor{red}{red} and spurious points (false positives) are in \textcolor{blue}{blue}.}
 \label{fig:qualitative_results}
\end{figure*}

We evaluate \textsc{SimpSyn} against the Synful baseline across various training and testing configurations. Table~\ref{tab:result_comparison} reports F1-scores for pre- and post-synaptic site detection across four EM datasets. Each cell shows the performance of \textsc{SimpSyn} (top) and Synful (bottom); diagonal entries (shaded) represent in-distribution (ID) performance, while off-diagonal entries reflect out-of-distribution (OOD) generalization. For each test domain, the best and second-best performing configurations are highlighted. See Supplementary Table S1 for a detailed synaptic pairwise comparison using F1 scores across all model configurations.

\textbf{In-distribution performance.} \textsc{SimpSyn} consistently outperforms Synful in all ID scenarios for both pre- and post-synaptic detection. On Hemibrain, \textsc{SimpSyn} achieves 0.783 (pre) and 0.606 (post) F1-scores—substantially higher than Synful's 0.262 and 0.437, respectively. Similarly, Octo and MANC models achieve strong ID performance, particularly in presynaptic detection (e.g., 0.538 on Octo vs. Synful's 0.150). These results suggest that \textsc{SimpSyn} handles intra-dataset variability and polyadic structures robustly, despite its architectural simplicity.

\textbf{Out-of-distribution generalization.} Generalization to unseen domains remains a challenging setting. Dataset-specific models show varying degrees of degradation under OOD shift. For example, a Hemibrain-trained model performs poorly on WASP (pre: 0.000, post: 0.000), likely due to differences in tissue contrast and synapse morphology. However, models trained on Octo or MANC overall transfer better to other domains, likely due to their structural diversity. Notably, the `All' model—trained jointly on all available training volumes achieves the highest average generalization performance across test domains, with strong pre-synaptic F1-scores (0.846, 0.654, 0.810, 0.736) and competitive post-synaptic results (0.635, 0.506, 0.664, 0.475). These results underscore the importance of anatomical and morphological diversity in training data.

\textbf{Asymmetries in detection.} Across datasets, we observe a general asymmetry between pre- and post-synaptic site performance. Presynaptic sites tend to be more reliably detected, while postsynaptic sites—especially in polyadic arrangements—show greater variability. This is visible both quantitatively in Table~\ref{tab:result_comparison} and qualitatively in Figure~\ref{fig:qualitative_results}. In regions with complex connectivity (e.g., WASP or MANC), \textsc{SimpSyn} predictions occasionally include spurious post sites or miss diffuse ones. This reflects both intrinsic biological ambiguity and architectural limitations in point-pair association. 

\textbf{Qualitative evaluation.} Figure~\ref{fig:qualitative_results} shows visual comparisons of \textsc{SimpSyn} predictions across datasets. True positives (green), false negatives (red), and false positives (blue) reveal several patterns. Hemibrain samples show relatively clean detections, while Octo and WASP images highlight the challenge of distinguishing faint synaptic sites from surrounding structures. Despite this, \textsc{SimpSyn} captures a large proportion of true sites without over-saturating predictions.
Overall, these results confirm that \textsc{SimpSyn}, despite being significantly simpler and smaller than Synful, offers competitive or superior accuracy across diverse domains. Its robustness across both ID and OOD conditions (when trained with representative data), with minimal tuning and reduced computational overhead, makes it a practical candidate for large-scale connectomics pipelines.

\subsubsection{Ablation Study of Post-processing Strategies}

\begin{table}[ht]
    \caption{\textbf{SimpSyn ablation experiments}. We conduct evaluations across all datasets and report the F1-scores for both pre- and post-synaptic site detection. Highest scores, and the default settings, are highlighted in \textbf{bold}.}
    \begin{subtable}[t]{0.485\textwidth}
    \subcaption{\textbf{Point creation method}. We explore two different approaches available in the scikit-image library~\cite{van2014scikit}.}
    \centering
    \begin{tabular}{lll}
    Method     & Pre  $\uparrow $  & Post $\uparrow $  \\ \midrule
    $peak\_local\_max$ & \textbf{0.762} & \textbf{0.570} \\
    $blob\_log$  & 0.675 & 0.482 \\
             &  &  \\
             &  &  \\
    \end{tabular}
    \label{subtab:ablationA}
    \end{subtable}%
    \hfill 
    \begin{subtable}[t]{0.485\linewidth}
    \subcaption{\textbf{Thesholding method}. An automatic way of thresholding can improve overall F1.}
    \centering
    \begin{tabular}{lll}
    Thresholding     & Pre $\uparrow$  & Post  $\uparrow $ \\ \midrule
    manual           & 0.664 & 0.464 \\
    auto             & 0.672 & 0.432 \\
    relative         & \textbf{0.762} & \textbf{0.570} \\
    relative (batch) & 0.717 & 0.533
    \end{tabular}
    \label{subtab:ablationB}
    \end{subtable}%
    \hfill 
    \begin{subtable}[t]{0.485\linewidth}
    \subcaption{\textbf{Point filtering}. Excluding detected pre synaptic sites that are in close proximity is justified, as such configurations are biologically implausible.}
    \centering
    \begin{tabular}{lll}
    Point filtering & Pre  $\uparrow $  & Post $\uparrow $  \\ \midrule
    No-filter          & 0.682 & \textbf{0.570}  \\
    By distance        & \textbf{0.762} & 0.558  \\
    By distance + mask        & 0.717 & 0.555 
    \end{tabular}
    \label{subtab:ablationC}
    \end{subtable}
\label{tab:ablation}
\end{table}

We conducted a set of ablations to assess how different post-processing strategies affect final detection performance. Table~\ref{tab:ablation} summarizes results under variations in point creation, thresholding, and filtering.

\textbf{Peak detection strategy:} As shown in Table~\ref{subtab:ablationA}, we compared $peak\_local\_max$ and $blob\_log$ from \texttt{scikit-image}~\cite{van2014scikit}. The former, which uses a straightforward local maximum criterion, outperforms $blob\_log$ in both precision and recall, while also requiring fewer hyperparameters and less compute. This method aligns well with our model’s spherical output and minimizes false mergers.

\textbf{Thresholding methods:} We evaluated manual, automatic, and relative thresholding strategies (Table~\ref{subtab:ablationB}). Relative thresholding consistently yielded the best results (pre: 0.762, post: 0.570), suggesting that global or per-batch intensity heuristics adapt well across volumes without requiring dataset-specific tuning.

\textbf{Distance-based filtering:} Synaptic sites predicted in close spatial proximity are often biologically implausible. Applying distance-based filtering improved precision, particularly for presynaptic sites (Table~\ref{subtab:ablationC}). Further constraining predictions using both distance and predicted masks reduced false positives, but also suppressed some valid detections—highlighting the trade-off between specificity and sensitivity.

These ablations reinforce a key design principle of \textsc{SimpSyn}: model simplicity can be complemented by targeted post-processing to achieve strong performance without relying on complex test-time inference pipelines. Moreover, when trained on anatomically diverse datasets, \textsc{SimpSyn} enables the discovery of generalizable post-processing heuristics that remain effective across invertebrate model organisms (see Supplementary Figure S1).
\section{Limitations}
While \textsc{SimpSyn} demonstrates strong in-distribution generalization and computational efficiency, several important limitations remain.

First, its out-of-distribution (OOD) generalization is sensitive to large domain shifts, particularly when models are trained on a single dataset. Variability in anatomical structure, polyadicity, and EM contrast continues to challenge detection robustness. Nonetheless, the model trained on the full cohort of datasets (`All') generalizes considerably better across held-out volumes, particularly for presynaptic sites. The lower performance observed for postsynaptic site detection can be partially attributed to the variability in synaptic partner distributions and configurations across different sub-volumes.

Second, although we report mean F1-scores across datasets, we do not currently include statistical confidence intervals or measures of variance. All \textsc{SimpSyn} models were trained with multiple random seeds, and we observed minimal performance fluctuations and is therefore not shown. However, a more rigorous treatment of variance and significance testing would strengthen future benchmarking, especially when evaluating small performance margins.


Third, we use nearest-neighbour strategy to associate a pre-synaptic site with multiple nearby post-synaptic partners in a polyadic synapse setting. This can lead to spurious post-site assignments—particularly in densely innervated regions, where spatial proximity alone is an unreliable indicator of true connectivity. As explicitly modelling polyadicity remains difficult~\cite{Buhmann2021AutomaticSet}, future work could explore contrastive objectives or uncertainty-aware assignment strategies to disambiguate true partners from nearby distractors.

Fourth, while \textsc{SimpSyn} is trained solely on sparse point annotations, offering clear efficiency advantages, it still depends on high-quality, manually curated ground truth. However, even expert annotators can introduce labelling inconsistencies, which can skew both model training and evaluation (see supplementary Fig S3). Addressing this issue may require human-in-the-loop proofreading, uncertainty-guided annotation, or robust objectives that account for label noise and incompleteness, especially in under-represented brain regions or species.


Finally, like other state-of-the-art approaches, \textsc{SimpSyn} is susceptible to spurious predictions. Given the minute scale of synaptic junctions and their visual similarity to nearby organelles, distinguishing true synapses from noise remains a non-trivial task. Incorporating contrastive training objectives or hard negative mining strategies may help improve specificity by explicitly modelling context and suppressing false positives.

\section{Conclusions and Future Work}
Accurate and generalizable synapse detection is a critical component of large-scale connectomic mapping, key for advancing our understanding of how structural connectivity underlies neural function in health and in disease. Here, we introduced \textsc{SimpSyn}, a lightweight, point-supervised model for synapse detection, and evaluated it across 16 connectomic volumes spanning multiple species and developmental stages. \textsc{SimpSyn} matches or surpasses a state-of-the-art baseline in both in-distribution and out-of-distribution scenarios. Ablation studies show that simple, biologically informed post-processing strategies—particularly relative thresholding and distance-based filtering—can substantially boost detection performance. Nevertheless, challenges remain in achieving robust out-of-distribution performance, particularly for postsynaptic site detection, due to polyadic variability and the lack of consistently prominent morphological features on the postsynaptic side. Future work will focus on enhancing pairing mechanisms and investigating label-denoising (uncertainty-aware) and contrastive learning strategies to further improve generalization within \textsc{SimpSyn}’s framework. These directions, combined with \textsc{SimpSyn}’s low computational overhead and reliance on sparse annotations, position \textsc{SimpSyn} as a promising tool for scalable and accessible connectome mapping pipelines. 

\section{Acknowledgments}
We thank Nicolo Ceffa (MRC Laboratory of Molecular Biology) for assistance with Blender rendering of mesh figures. We are grateful to Ana Correia, Amina Dulac, Michael Clayton, and Samuel Harris for proofreading synapses in the Octo dataset.

\section{Funding}
This project was funded by a Wellcome Trust Investigator award to AC (Ref: 205038/Z/16/Z), ERC (Ref: ERC-2018-COG: 819650), BBSRC (Ref: APP26929, Population Connectomics) and the MRC LMB core funding.

\section{Author contributions}
The author contributions are as follows: 

Conceptualization: SM and DFB; Methodology: SM and DFB; Investigation: SM and DFB; Visualization: SM, DFB, SYL and AC; Supervision: AC; Writing—original draft: SM and DFB; Writing—review and editing: SM, DFB, SYL, and AC

\section{Competing Interests}
The authors declare no competing interests.

\section{Data and Materials Availability}
All data needed to evaluate the conclusions are provided in the paper and/or the Supplementary Materials. Curated EM datasets will be released on Zenodo (DOIs to be added upon publication). Source code is available at \href{https://github.com/Mohinta2892/catena/tree/dev}{Catena} and \href{https://github.com/BiaPyX/BiaPy}{BiaPy}.

\bibliography{arxiv_2025}

\clearpage


\section*{Supplementary material}

# Towards Generalized Synapse Detection Across Invertebrate Species
## Supplementary Material

Table S1: **Synaptic pairwise comparison** measured using F1 scores. The upper row denotes the test datasets, while the first column indicates the training dataset. Consequently, diagonal entries (in gray) correspond to in-distribution (ID) evaluations, whereas off-diagonal entries represent out-of-distribution (OOD) scenarios. Within each cell, the upper value refers to performance achieved by SIMPSYN, and the lower value to that of Synful. For each column, the highest score is highlighted in **bold**, and the second-highest is underlined.

| Train \ Test | Hemibrain | Octo | WASP | MANC | All |
|---|---|---|---|---|---|
| Hemibrain | **0.832** / 0.401 | 0.014 / 0.155 | 0.0 / 0.0 | 0.021 / 0.142 | 0.217 / 0.175 |
| Octo | 0.194 / 0.157 | 0.618 / 0.320 | 0.003 / 0.196 | 0.215 / 0.176 | 0.258 / 0.212 |
| WASP | 0.006 / 0.0 | 0.318 / 0.032 | <u>0.901</u> / 0.766 | 0.574 / 0.651 | 0.450 / 0.362 |
| MANC | 0.032 / 0.007 | 0.522 / 0.0 | 0.0 / 0.109 | 0.680 / 0.109 | 0.310 / 0.056 |
| All | <u>0.687</u> / 0.385 | **0.624** / <u>0.620</u> | **0.909** / 0.875 | <u>0.719</u> / **0.732** | **0.734** / <u>0.653</u> |

Table S2: **Synful MT1 Model Hyperparameters**

| Parameter | Value |
|---|---|
| Input Size | $128 \times 128 \times 128$ |
| Downsample Factors | $[2,2,2] \to [2,2,2] \to [2,2,2]$ |
| Feature Maps (fmap_num) | 12 |
| Feature Map Increment (fmap_inc_factor) | 5 |
| Losses | BCE (mask) and MSE (vector) |
| Learning Rate | $5 \times 10^{-5}$ |
| Loss Combination Type | `sum` |
| Main Loss Scale (m_loss_scale) | 1.0 |
| Direction Loss Scale (d_loss_scale) | 1.0 |
| Reject Probability | 0.95 |
| Blob Radius | 20 |
| Max Iterations | 300,000 |
| Blob Mode | `ball` |
| Direction Scale (d_scale) | 1 |
| Direction Blob Radius (d_blob_radius) | 150 |
| Clip Range | $[7 \times 10^{-4}, 0.9993]$ |
| Voxel Size | $8 \times 8 \times 8$ |

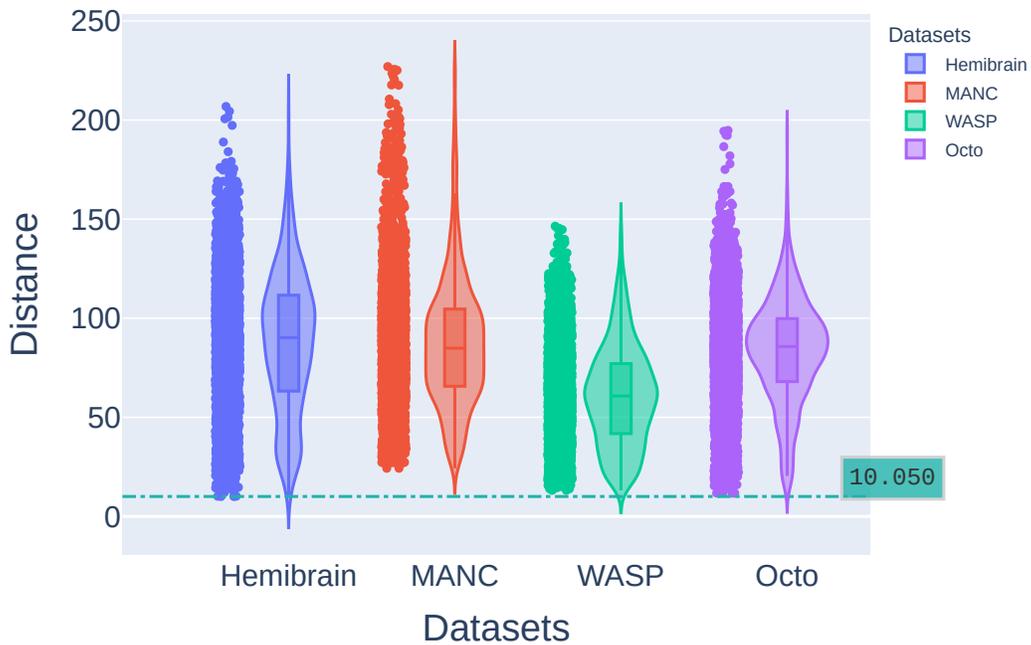

Figure S1: **Pre point distribution in the ground truth data across all datasets**. The green line and its corresponding value in the annotation 10.050 represent the minimum distance (in pixels) found across all datasets. This translates to $80nm$ taking into account that the data resolution is $8 \times 8 \times 8$ for all datasets.

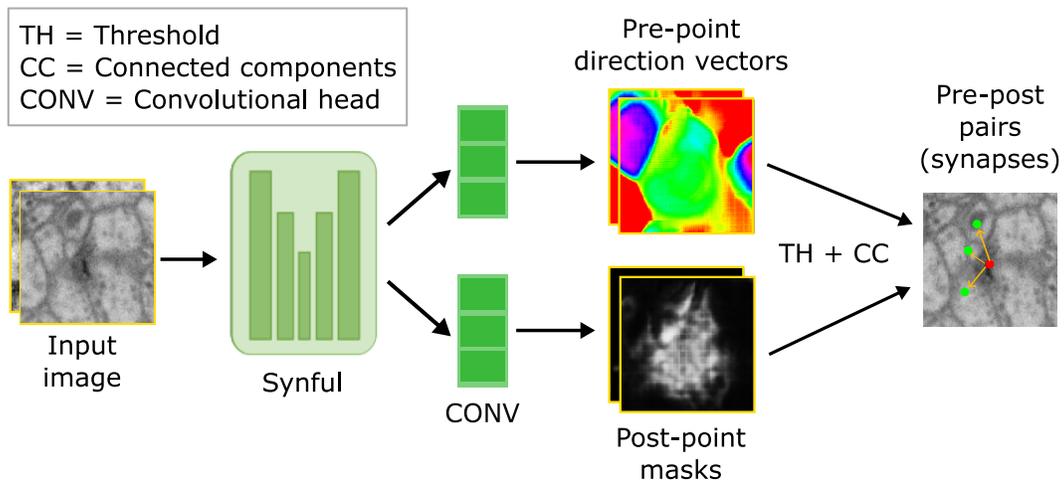

Figure S2: **Synful schematic**. Our Synful baseline uses a double-headed 3D unet architecture (MT1), which jointly predicts the post-synaptic masks and the pre-synaptic direction vectors. Final pre-post synaptic pairs are extracted after thresholding and grouping connected components.

Table S3: **Hyperparameter search for SIMPSYN, exploring the effect of input field of view on F1-score**. Larger patch sizes without downsampling along the Z-axis across all four levels of the Residual U-Net lead to improved performance.

| Patch size | Z down. | Feature maps | Pre ↑ | Post ↑ |
| --- | --- | --- | --- | --- |
| (128, 128, 128, 1) | ✗ | [16,32,64,128,256] | 0.721 | 0.557 |
| (64, 64, 64, 1) | ✗ | [16,32,64,128,256] | 0.717 | 0.528 |
| (64, 128, 128, 1) | ✗ | [16,32,64,128,256] | 0.72 | 0.556 |
| (80, 80, 80, 1) | ✗ | [32, 64, 128, 256] | 0.711 | 0.546 |
| (80, 80, 80, 1) | ✓ | [32, 64, 128, 256] | 0.706 | 0.546 |
| (128, 128, 128, 1) | ✓ | [32, 64, 128, 256] | 0.738 | 0.564 |
| (128, 128, 128, 1) | ✗ | [32, 64, 128, 256] | **0.762** | **0.570** |

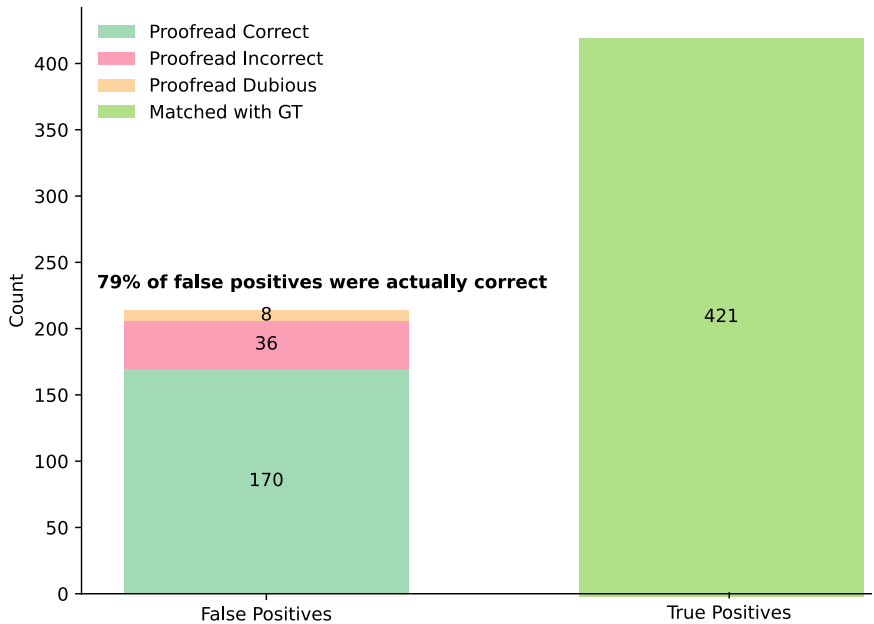

Figure S3: **Breakdown of synapse detection outcomes and proofreading results**. The left bar shows the composition of model-predicted 214 false positives in Octo. After manual proofreading by a domain expert : 170 (79%) were found to be valid synapses missed in the ground truth, while 36 were incorrect and 8 remained ambiguous. The right bar shows ground truth–matched true positives. These results underscore the inherent complexity of EM data and the resulting annotation incompleteness, which can lead to the penalization of correct predictions during evaluation.



\end{document}